\documentclass[runningheads]{llncs}

\usepackage{eccv}

\usepackage{eccvabbrv}

\usepackage{graphicx}
\usepackage{booktabs}

\usepackage[accsupp]{axessibility}  % Improves PDF readability for those with disabilities.

\usepackage{hyperref}

\usepackage{orcidlink}

\begin{document}

\title{Evaluation Framework for Feedback Generation Methods in Skeletal Movement Assessment} 

\titlerunning{Evaluation Framework for Feedback Generation Methods}

%\author{Tal Hakim\inst{1}\orcidlink{0009-0009-3038-2621} }
\author{Tal Hakim\orcidlink{0009-0009-3038-2621} }

\authorrunning{T. Hakim}

\institute{SMARTSHOOTER\\
\email{thakim.research@gmail.com}
}

\maketitle

\begin{abstract}
  The application of machine-learning solutions to movement assessment from skeleton videos has attracted significant research attention in recent years. This advancement has made rehabilitation at home more accessible, utilizing movement assessment algorithms that can operate on affordable equipment for human pose detection and analysis from 2D or 3D videos. While the primary objective of automatic assessment tasks is to score movements, the automatic generation of feedback highlighting key movement issues has the potential to significantly enhance and accelerate the rehabilitation process. While numerous research works exist in the field of automatic movement assessment, only a handful address feedback generation. In this study, we propose terminology and criteria for the classification, evaluation, and comparison of feedback generation solutions. We discuss the challenges associated with each feedback generation approach and use our proposed criteria to classify existing solutions. To our knowledge, this is the first work that formulates feedback generation in skeletal movement assessment.
\end{abstract}

\section{Introduction}
\label{sec:intro}
Research into analyzing human movements through skeletal videos has seen significant growth in recent years~\cite{han2017space}. Such analysis serves various purposes, including action recognition~\cite{ohn2013joint}, person identification~\cite{kviatkovsky2015person}, prediction of medical conditions~\cite{masalha2020predicting,eichler2022automatic}, and movement assessment~\cite{sardari2023artificial,frangoudes2022assessing,da2015motor,maudsley2017comparative,debnath2022review,hakim2020comprehensive}. Importantly, these analyses often play a crucial role in monitoring and supporting rehabilitation processes~\cite{deb2022graph,hakim2019mal,kryeem2023personalized,eichler20183d,eichler2018non}.

Performing medical assessments using AI-based systems offers the advantage of patients being able to conduct assessments in their own homes. Additionally, when monitoring the rehabilitation process, maintaining consistency in assessment criteria is facilitated, as scores are generated consistently by the same system. This stands in contrast to in-person assessments, where professionals, each with their own judgment emphases, may vary over time.

The primary drawback of automatic assessment at home lies in providing feedback and guidance for improvement. Similar to how loss-gradient guides iterative optimization algorithms by indicating direction, feedback plays a crucial role in guiding patients towards improved performance over time. In face-to-face assessments, professionals can offer direct guidance to patients between or even during performances. However, in an automatic assessment system, providing such guidance is not as straightforward or immediate. It may require the utilization of advanced machine-learning techniques and human-computer interaction practices.

While numerous research works exist in the field of automatic movement assessment, only a handful provide algorithms for feedback generation. This paper aims to address this gap by examining the nature and significance of feedback within the context of movement assessment. Specifically, we will explore the types of feedback that can be generated, discuss the approaches and challenges involved in feedback generation, and use the proposed terminology to classify existing solutions in this area.

This work addresses feedback generation as a computational problem and does not cover the deployment process in clinical scenarios.

\section{Movement Assessment}
Movement assessment, as one of the applications of movement analysis from skeletal videos, presents a relatively challenging task. Unlike action recognition, which typically involves classifying movements from a high-level temporal and skeletal perspective, movement assessment requires recognizing low-level details crucial for performance quality and closely attending to them at specific stages of movement progress~\cite{hakim2020comprehensive}. Automatic movement assessment solutions find application across various domains, including rehabilitative medical assessments~\cite{su2013personal,hakim2019mal,eichler2018non,masalha2020predicting,kryeem2023personalized,osgouei2018objective,cao2019novel,williams2019assessment,capecci2018hidden,zhang2020rehabilitation,liao2020deep,ding2023design,lee2020towards,kanade2022robust,osgouei2020rehabilitation}, gymnastics~\cite{parisi2016human,bi2022lazier,hulsmann2018classification,lin2013kinect}, physical skills~\cite{dressler2020towards,hu2014real,yu2019dynamic}, and Olympic sports~\cite{lei2020learning}.

Fundamental components of general movement analysis solutions comprise skeleton detection, geometric normalization, and feature extraction. Movement assessment solutions typically incorporate temporal alignment and score prediction, treated as either classification or regression tasks~\cite{hakim2020comprehensive}. Therefore, the primary objective of movement assessment solutions is to predict performance scores for specific movements on predefined scales, which can be continuous or discrete. Ground-truth for training the score prediction model is typically supplied by professionals. Some models are trained on samples with labels distributed across the entire score scale, while others are trained solely on proper performances~\cite{hakim2020comprehensive}.

Feedback generation, which holds the potential to accelerate movement learning and, consequently, the rehabilitation process~\cite{sardari2023artificial,frangoudes2022assessing,hakim2020comprehensive,debnath2022review,cunha2023home}, is occasionally addressed by studies focusing on movement assessment~\cite{benettazzo2015low,hakim2019mal,kryeem2023personalized,lee2020towards,parisi2016human,hulsmann2018classification,leightley2016automated}. Feedback is sometimes equated with interpretability, which entails understanding why a specific outcome is observed and how input changes will alter it~\cite{frangoudes2022assessing}. In other cases, it is referred to as guidance~\cite{garg2023short}. As we discuss in this paper, the concept of feedback generation can be interpreted in multiple ways. Evaluating the quality of generated feedback can be challenging. From a data perspective, annotating training and test samples with their expected feedback may be considerably more complex than solely annotating them with their expected performance scores. Moreover, in certain scenarios, multiple feedback outputs may be deemed correct or partially correct, posing challenges for evaluation. In the subsequent sections of this paper, we delve deeper into feedback generation.

\section{Feedback Generation}
Defining ``feedback'' can be complex. In the context of movement assessment, it encompasses a wide array of response types, occurring during or after a performance. Feedback may consist of both basic and advanced guidance, conveyed verbally~\cite{benettazzo2015low,hulsmann2018classification,lee2020towards}, textually~\cite{antunes2016visual,hakim2019mal,lee2020towards,kryeem2023personalized,garg2023short}, or visually~\cite{benettazzo2015low,antunes2016visual,devanne2016learning,parisi2016human,hulsmann2018classification,mourchid2023d}, and occasionally through physical means~\cite{galeano2014tool}. Table~\ref{tbl:review} provides a summary of feedback generation solutions, based on the criteria proposed in the following sections.

\subsection{Scoring Model} The primary objective in movement assessment is score generation. Typically, scoring is accomplished through regression or classification machine-learning algorithms, trained with sample data to minimize prediction errors~\cite{hakim2020comprehensive}. These algorithms often derive compound features from various subsets of raw geometric pose and movement features, or handcrafted features, aiding in prediction accuracy. However, interpreting compound features can be challenging, making it less immediate to identify performance errors and generate corresponding feedback. Alternatively, models that solely utilize raw features may offer clearer interpretability, enabling a more straightforward identification of feature deviations that have the highest impact on the score, which can then be translated into feedback~\cite{hakim2019mal}. For instance, a single component of a single 3D joint location vector, or velocity vector, may be a raw feature. For example, a deviation of the Y component of one of the elbows can be translated directly to feedback regarding the elbow's height. The raw features approach may be more adaptive in assessing and providing feedback for new movement types without prior assumptions, due to their general nature. In cases where using raw features is not feasible, another statistical model may be necessary to predict or detect mistakes and translate them into feedback~\cite{kryeem2023personalized,parisi2016human}.

\subsection{Feedback Boundness and Order} 
\begin{figure}[t]
\centering
  \includegraphics[width=\linewidth]{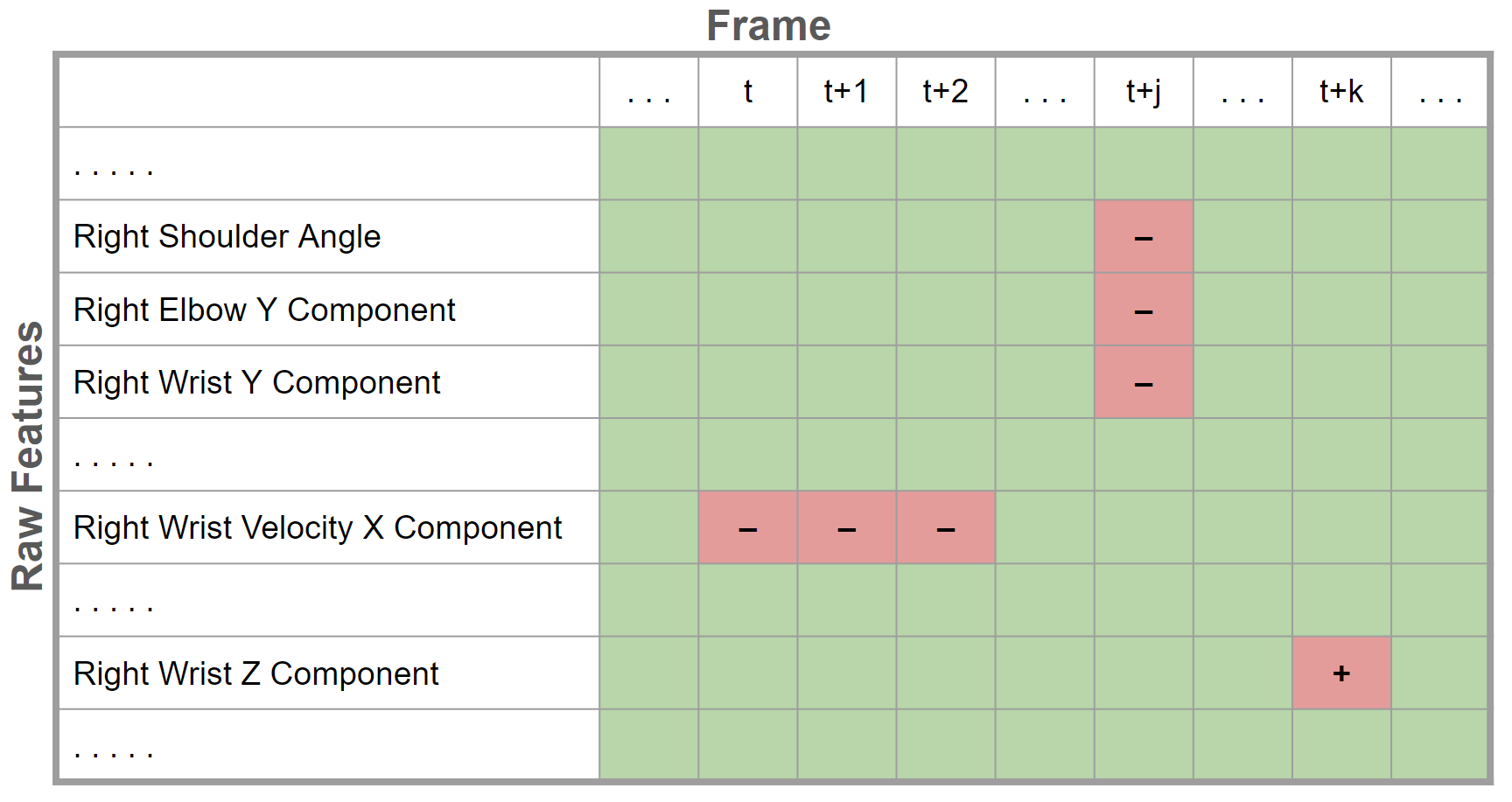}
  \caption[]{Examples of raw feature deviations. A green, clear cell indicates a proper, in-range feature value, while cells highlighted in red indicate deviating feature values. Each deviation can be directly translated into a first-order feedback label. When a feature consistently deviates over a sequence of frames, temporal aggregation will consolidate the sequence of first-order feedback labels into a single feedback label. Similarly, when related features deviate simultaneously, skeletal aggregation will consolidate the set of related first-order feedback labels into a single feedback label.}
  \label{fig:firstOrderDeviations}
\end{figure}

According to \cite{hakim2020comprehensive}, feedback can be categorized as either bound or unbound. Feedback is considered bound when it is limited to predefined mistakes that can be detected~\cite{kryeem2023personalized,parisi2016human}. Conversely, unbound feedback may encompass detected and interpreted mistakes, or deviations, in any provided raw feature, translating them into text or visual representation~\cite{hakim2019malarxiv}. Another parameter used to classify feedback types is the feedback order. First-order feedback labels refer to mistakes, or deviations, related to a single raw feature at a specific point in time. Second and third-order feedback labels may involve temporal and skeletal aggregations, resulting in more abstract and intuitive output. When only one of the two aggregation types is applied, the feedback labels will be considered second-order, whereas when both are applied, they will be considered third-order. Raw feature deviations are demonstrated in Figure~\ref{fig:firstOrderDeviations}. The advantage of bound feedback lies in the ability of the prediction model to be trained directly to predict higher-order feedback~\cite{kryeem2023personalized,parisi2016human}. This is possible by addressing the bound feedback generation problem as a detection or classification problem, where the set of predefined labels can include abstract, high-order labels. Such labels, for example, can include information regarding balance, strength, number of repetitions, or other details that may not be easily derived from raw feature deviations. Meanwhile, with unbound feedback, the aggregation step from first-order feedback to higher-order feedback is not always straightforward and may require the design of a problem formulation, which may be addressed using heuristics or optimization problems~\cite{hakim2019mal}. Recent advancements in Natural Language Processing (NLP), such as Large Language Models (LLMs)~\cite{vaswani2017attention,devlin2018bert}, including tools like ChatGPT, might allow an automated feedback aggregation, which will not be constrained by any proposed formulation.

\subsection{Temporal Granularity} 
\begin{figure}[t]
\centering
  \includegraphics[width=\linewidth]{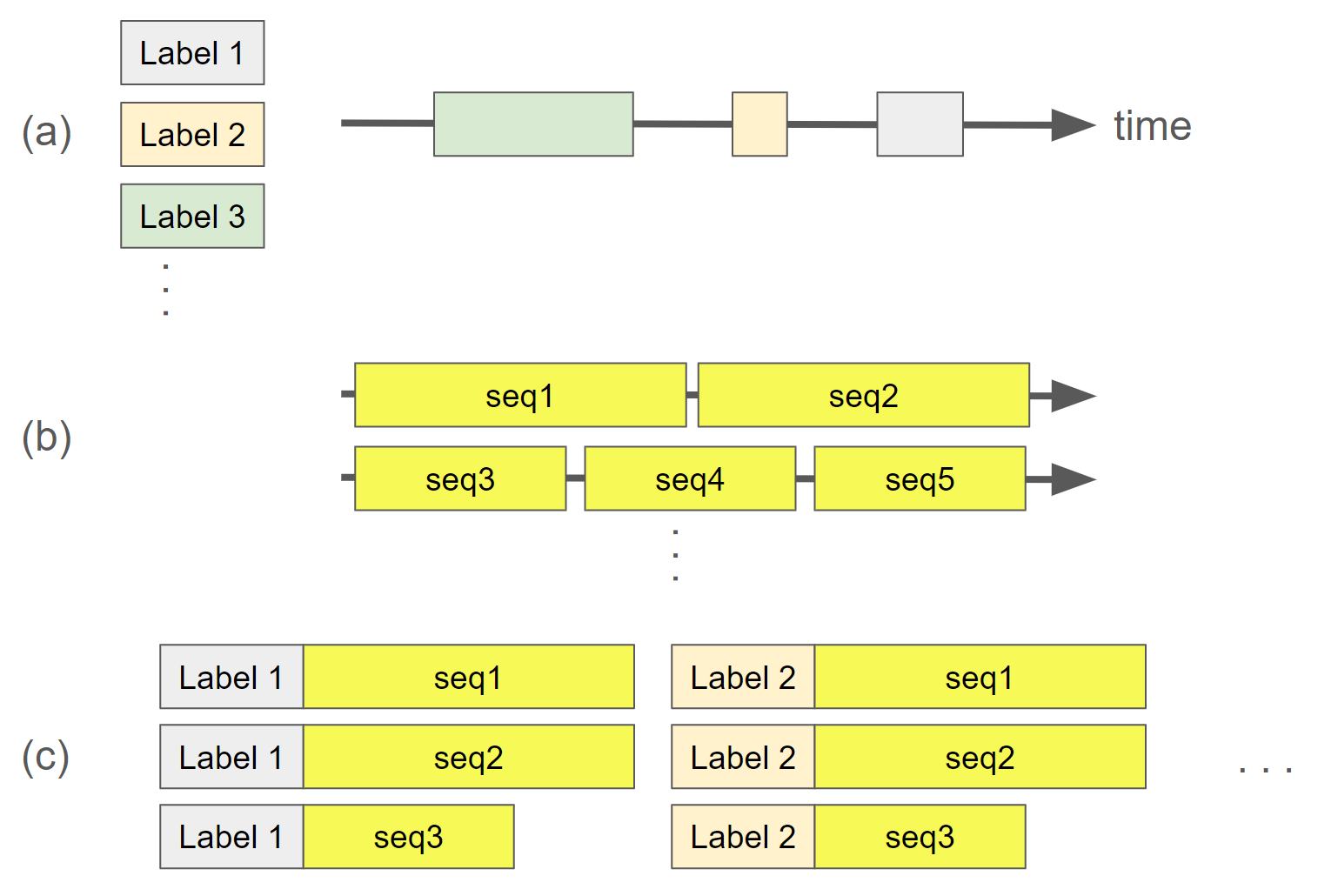}
  \caption[]{Different approaches to adding temporal notation to bound feedback labels: (a) different feedback labels are detected along the movement time axis using sliding windows of different lengths; (b) a set of time frames is defined to support temporally local feedback generation; and (c) temporally local feedback labels are achieved using a Cartesian product of the sets of feedback labels and time frames.}
  \label{fig:temporalDetection}
\end{figure}

Generated feedback may comprise local mistakes detected along the temporal dimension or feedback labels referring to the entire movement, termed as temporally global feedback. The prediction of temporally global feedback can be achieved by employing a set of temporally global feedback label classifiers on a representation of the entire movement or by employing heuristics or temporal aggregations on locally detected mistakes. Temporally local mistakes, or feedback, can be predicted as either bound or unbound feedback. In the case of unbound feedback, unless aggregated over the entire temporal dimension, the feedback remains temporally local by definition, whereas for bound feedback, temporally local feedback can be predicted using either temporally sliding window classification or by expanding the set of feedback labels to include temporal localization. This expansion can be achieved by multiplying the set of feedback labels by a set of time frames, as illustrated in Figure~\ref{fig:temporalDetection}.

\subsection{Offline/Online Feedback Generation Mode} In the domain of automatic movement assessment, movements are represented as sequences of skeleton poses, inherently possessing a temporal dimension. Recognizing the stages of movement progression during assessment, termed as temporal alignment, is crucial due to variations in movement progress across different performances~\cite{hakim2020comprehensive}. Some studies employ online alignment techniques using tools like Hidden Markov Models (HMMs)~\cite{paiement2014online,devanne2016learning} and Recurrent Neural Networks (RNNs)~\cite{jun2020feature,cao2019novel,nguyen2018estimating,nguyen2018skeleton,parisi2016human}, while others opt for offline methods such as Dynamic Time Warping (DTW)~\cite{chaaraoui2015abnormal,su2013personal,dressler2019data,dressler2020towards,yu2019dynamic}, piecewise warping between temporal points-of-interest~\cite{hakim2019mal}, and DTAN~\cite{shapira2019diffeomorphic,kryeem2023personalized}. Offline temporal alignment poses a challenge for generating online feedback during performance analysis, as it only initiates after the entire performance has concluded. Conversely, online temporal alignment allows for ongoing analysis after each frame input, potentially enabling the production of online feedback. This capability holds promise in bridging one of the significant gaps between automatic and frontal assessments.

\subsection{Feedback Modality} 
\begin{figure}[t]
\centering
  \includegraphics[width=\linewidth]{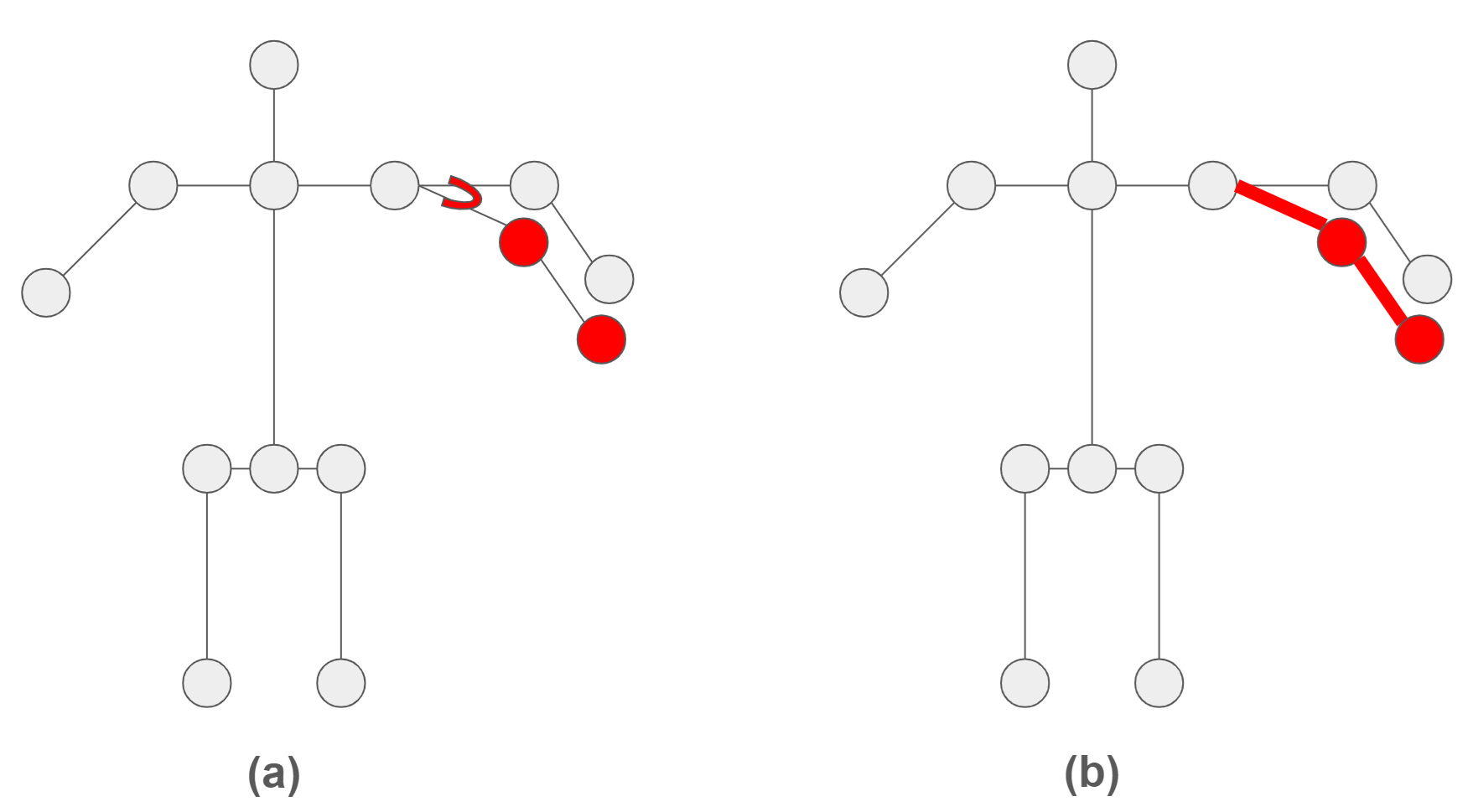}
  \caption[]{Visualizing multiple first-order feedback labels in (a), versus a single second-order feedback label in (b). This example demonstrates the power of visual feedback over textual feedback: the three reported deviations in (a) are first-order, yet, they provide intuitive feedback, leaving negligible added value for the aggregated feedback label in (b). This approach has the potential to enable unbound feedback generation without addressing the skeletal aggregation problem.}
  \label{fig:visualFeedback}
\end{figure}

Creating textual feedback that is both intuitive and informative presents challenges, as discussed earlier. In most cases, verbal feedback will be derived from textual feedback using text-to-speech technologies, and thus it encounters similar challenges. In contrast, visual feedback holds the potential to be more intuitive and informative, and sometimes even easier to generate. For instance, presenting multiple first-order unbound feedback labels simultaneously as an animation on a skeleton figure inherently incorporates skeletal and temporal aggregation, as illustrated in Figure~\ref{fig:visualFeedback}.

\section{Evaluating Generated Feedback}
Typically, evaluating a prediction task involves comparing it to a provided test dataset with labels, also known as ground-truth. With feedback varying in orders, whether bound or unbound, temporally global or local, online or offline, defining feedback ground-truth can vary, leading to multiple ways to assess the quality of generated feedback.

When predicting unbound feedback, the aggregation algorithms that elevate the feedback order may significantly influence the quality evaluation of the generated feedback. Similarly, heuristics and temporal aggregation algorithms that transform temporally local feedback labels into global ones may also play a crucial role in this evaluation.

\subsection{Bound Feedback Labels} In scenarios where feedback labels are bound, the evaluation task is relatively straightforward. In such cases, the ground-truth feedback labels belong to a predefined set of labels. If not all labels are temporally global, temporal notations must be included in the labels. As discussed in the previous section, implementing temporal notation for feedback labels can be achieved by either incorporating a temporal bounding box property or inherently including sufficient temporal information in the set of feedback labels. During evaluation, a predefined temporal Intersection over Union (IoU) threshold is used as a criterion to consider a generated feedback label as temporally correct. The primary advantage of using bound feedback labels lies in the annotation process, which only requires professionals to specify labels from a predefined set for each sample. However, in such cases, it may be necessary to train the feedback generation model on numerous samples of each feedback label to learn the association between movement features and ground-truth feedback labels.

\subsection{Unbound Feedback Labels} Similar to generated unbound feedback, the ground-truth labels for unbound feedback may require inclusion of the set of deviated features, their corresponding types of deviations, and a temporal bounding box. When ground-truth feedback labels lack skeletal aggregation, each feature and its deviation are denoted as separate feedback labels. Conversely, when feedback labels include skeletal aggregation, all involved features are encompassed within a single label. Unlike bound feedback labels, annotating unbound feedback labels can be an exhaustive task, as each feature must be individually assessed by a professional, considering aggregations as well. However, these annotations may solely be utilized for feedback evaluation, as unbound feedback is typically generated from aggregated first-order deviations, which do not necessarily require training on feedback labels.

\subsection{Feature IoU} In the case of unbound feedback generation, temporal IoU is a common means for evaluating the temporal accuracy of temporally-aggregated feedback labels. Likewise, when applying skeletal aggregation, feature IoU should be utilized to measure the quality of generated feedback labels. This involves evaluating the similarity between the predicted and ground-truth sets of involved features and their deviations. Specifically, this entails calculating a discrete IoU between two sets of deviated features, denoted as $P$ and $G$, representing the predicted and ground-truth deviated features, respectively. Each deviated feature consists of a tuple indicating the feature identity and the deviation type. Essentially, this calculation represents the Jaccard similarity between the two sets,

\begin{equation}
    J(P,G) = \frac{|P \cap G|}{|P \cup G|}.
\end{equation}

\subsection{Online-Feedback Evaluation} Evaluating feedback generated before a performance has been completed can be challenging from multiple aspects. From the ground-truth perspective, the feedback ground-truth for one performance could differ between considering the entire performance versus just a section of it. This limitation should be considered when defining the annotation format, potentially complicating the annotation process. From the evaluation perspective, a process should be established to assess the accuracy of all generated feedback collectively. For instance, this could involve calculating the mean accuracy of generated feedback across all frames where feedback was expected.

\begin{table*}[]
\centering
\caption[]{Existing works dealing with feedback generation. In Evaluation Type, (Q) stands for quantized (bound) feedback labels. Feedback order is only reported for unbound feedback in textual modality.}

\resizebox{1.0\linewidth}{!}{%
\begin{tabular}{l|c||c|c|c|c|c|c}
\hline
\hline
\textbf{Ref.} & \textbf{Year} & \textbf{Movement Domain} & \textbf{Modalities} & \textbf{Boundness, Order} & \textbf{Gen. Mode} & \textbf{Temp. Granul.} & \textbf{Evaluation Type} \\
 \hline
 \cite{benettazzo2015low} & 2015 & Physical Rehabilitation & Visual,Verbal & Bound & Online & Local & N/A \\
\hline
\cite{devanne2016learning} & 2016 & Gait & Visual & Unbound & Online & Local & Qualitative \\ 
\hline
\cite{parisi2016human} & 2016 & Powerlifting & Visual & Unbound & Online & Local & Quantitative(Q) \\
\hline
\cite{antunes2016visual,baptista2017video} & 2016-17 & Stroke Rehabilitation & Text,Visual & Unbound (order=3) & Offline & Local,Global & Qualitative \\
\hline
\cite{hulsmann2018classification} & 2018 & Squats and Tai Chi & Visual,Verbal & Unbound & Online & Local & N/A \\
\hline
\cite{hakim2019mal,hakim2019malarxiv} & 2019 & Stroke Rehabilitation & Text & Unbound (order=2) & Offline & Local,Global & Qualitative \\
\hline
\cite{lee2020towards} & 2020 & Stroke Rehabilitation & Text,Verbal & Unbound (order=N/A) & Online & N/A & N/A \\
\hline
\cite{kryeem2023personalized} & 2023 & Hip-Replacement Recovery & Text & Bound & Offline & Global & Quantitative \\
\hline
\cite{mourchid2023d} & 2023 & Physical Rehabilitation & Visual & Unbound & Online & Local & Qualitative \\
\hline
\cite{garg2023short} & 2023 & Physical Rehabilitation & Text & Unbound (order=1) & Offline & N/A & Quantitative(Q) \\
\hline
\end{tabular}
}
\label{tbl:review}
\end{table*}

\section{Discussion}
In this study, we aimed to explore the definition of feedback within the context of automatic movement assessment, along with its various forms and the challenges associated with providing solutions for each. We have addressed several aspects in which solutions may vary, with particular emphasis on the concept of feedback boundness.

In terms of feedback boundness, bound feedback offers several advantages, including a relatively straightforward annotation process, the ability to utilize off-the-shelf machine learning tools such as classification and detection for building feedback solutions, and the inherent generation of high-order feedback as predefined labels. However, bound feedback is limited in its scope compared to unbound feedback, faces challenges in producing temporally local feedback, and may require a substantial number of annotated training samples for each feedback label. On the other hand, unbound feedback offers the advantage of greater flexibility, potentially supporting scenarios not initially considered during system design. Additionally, unbound feedback generation algorithms may not require training on feedback samples, as they are based on first-order deviations from a proper performance, inherently providing temporally local feedback. However, unbound feedback poses challenges in providing high-order feedback and may require exhaustive annotation of feedback labels not bound to a predefined dictionary for evaluation.

Another orthogonal aspect we have discussed is online versus offline feedback generation. While online feedback generation holds the potential to accelerate the rehabilitation process and enhance the overall experience, evaluating its accuracy may pose challenges, both in terms of annotation and providing a comprehensive accuracy metric.

Another aspect we have explored is the feedback modality. We have proposed that, especially in cases where feedback is unbound, generating textual feedback without overwhelming the recipient might be challenging. In contrast, when presented visually, even raw, first-order feedback may be interpreted more intuitively.

In upcoming studies focusing on feedback generation, adopting the terminology and  criteria outlined in this paper can be beneficial. For instance, in a proposed solution generating bound feedback, it is beneficial to mention when the feedback labels are of high order and if a quantitative evaluation is included. Furthermore, acknowledging the need for numerous training samples for each feedback label might justify a relatively limited number of supported feedback labels. Similarly, in a proposed solution generating unbound feedback, it is valuable to underscore feedback versatility while mentioning the impracticality of annotating feedback labels, which might justify an inclusion of qualitative evaluation only.

\section{Conclusions}
This study addresses the critical gap in automatic movement assessment by focusing on the generation of feedback, which is essential for enhancing and accelerating the rehabilitation process. By proposing a clear terminology and criteria for the classification, evaluation, and comparison of feedback generation solutions, we provide a structured approach to tackling the challenges inherent in this area. Our classification framework helps in understanding and improving existing solutions, promoting the development of more effective and interpretable feedback mechanisms. As the first work to formulate feedback generation in skeletal movement assessment, our research lays the groundwork for future advancements, aiming to make rehabilitation more accessible and efficient through sophisticated machine-learning algorithms.

\section*{Acknowledgement}
I dedicate this work to our beloved dog, Benji, who sadly passed away while this study was in progress, at the age of 16 years and 8 months. Benji had a pure soul and was a true friend, playing a significant role in our lives, particularly in mine. He was always by my side, providing companionship and support throughout much of my career journey. Unfortunately, during his final month, he battled heart and liver disease along with other health issues, compromising his quality of life. Despite our efforts to aid his recovery, it became evident that his suffering was too great, and we made the difficult decision to euthanize him. My dear Benji, losing you after all these years together has been so traumatic. The realization that you are no longer around and that I will never see you again  is heartbreaking. I miss you. I love you and I will never forget you. May your soul rest in peace.

\begin{center}
\includegraphics[width=\linewidth]{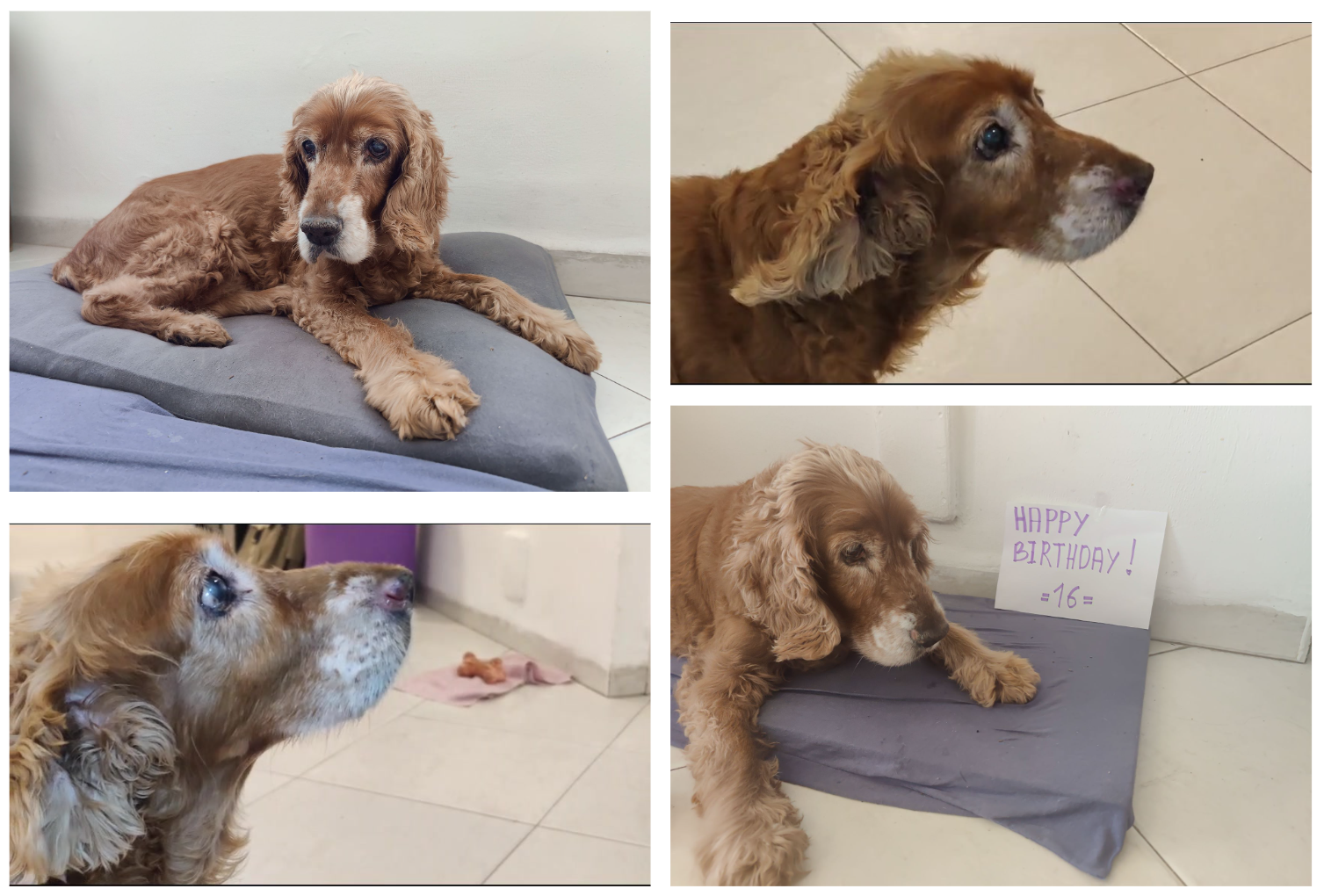}
\end{center}

\bibliographystyle{splncs04}
\bibliography{main}
\end{document}